\def\year{2021}\relax
\documentclass[letterpaper]{article} % DO NOT CHANGE THIS
\usepackage[switch]{lineno}
\usepackage{aaai21}  % DO NOT CHANGE THIS
\usepackage{times}  % DO NOT CHANGE THIS
\usepackage{helvet} % DO NOT CHANGE THIS
\usepackage{courier}  % DO NOT CHANGE THIS
\usepackage[hyphens]{url}  % DO NOT CHANGE THIS
\usepackage{graphicx} % DO NOT CHANGE THIS
\usepackage{natbib}  % DO NOT CHANGE THIS AND DO NOT ADD ANY OPTIONS TO IT
\usepackage{caption} % DO NOT CHANGE THIS AND DO NOT ADD ANY OPTIONS TO IT
\usepackage[utf8]{inputenc} % allow utf-8 input
\usepackage{url}            % simple URL typesetting
\usepackage{booktabs}       % professional-quality tables
\usepackage{amsfonts}       % blackboard math symbols
\usepackage{nicefrac}       % compact symbols for 1/2, etc.
\usepackage{microtype}      % microtypography
\newsavebox{\measurebox}

\usepackage{bm}
\usepackage{soul}
\usepackage{amsmath}
\usepackage{amsthm}
\usepackage{bbding} % for the \CheckmarkBold symbol
\usepackage{xcolor}
\usepackage{enumitem}
\usepackage{subcaption}
\usepackage{algpseudocode}
\usepackage{multirow}
\usepackage{breqn}

\title{Improving Aerial Instance Segmentation in the Dark with \\
Self-Supervised Low Light Enhancement}
\author {
        Prateek Garg,\textsuperscript{\rm 1}
        Murari Mandal,\textsuperscript{\rm 2}
        Pratik Narang\textsuperscript{\rm 3} \\
}
\affiliations {
    \textsuperscript{\rm 1}Department of Electrical and Electronics Engineering, BITS Pilani, Pilani, Rajasthan, India\\
    \textsuperscript{\rm 2}Department of Computer Science and Engineering, IIIT Kota, Rajasthan, India \\
    \textsuperscript{\rm 3}Department of Computer Science and Information Systems, BITS Pilani, Pilani, Rajasthan, India \\
    f20180412@pilani.bits-pilani.ac.in,
    murarimandal.cv@gmail.com, pratik.narang@pilani.bits-pilani.ac.in \\
}

\begin{document}
\maketitle
\begin{abstract}
Low light conditions in aerial images adversely affect the performance of several vision based applications. There is a need for methods that can efficiently remove the low light attributes and assist in the performance of key vision tasks. In this work, we propose a new method that is capable of enhancing the low light image in a self-supervised fashion, and sequentially apply detection and segmentation tasks in an end-to-end manner. The proposed method occupies a very small overhead in terms of memory and computational power over the original algorithm and delivers superior results. Additionally, we propose the generation of a new low light aerial dataset using GANs, which can be used to evaluate vision based networks for similar adverse conditions.
\end{abstract}

\section{Introduction} 
Most of the computer vision methods are biased towards highly clean and sanitised dataset, which is hardly the scenario in most practical settings. The data on which these methods are trained lack substantially in adverse aspects such as low lighting conditions and unwanted noise, which demand immediate attention if the methods are to be utilised in real-time. Moreover, the methods which are being developed do not include any distinctive mechanism to deal with such complications, hence supplementing the underlying problem. 

Amongst many adverse problems, low light conditions in aerial imagery is a prominent one. Low light is an inevitable part of aerial images, since they may be captured with insufficient light due to extreme weather constraints, night time, poor capturing techniques, low contrast conditions, inept position of ground objects etc. 
This makes it very difficult to accommodate key UAV applications such as remote sensing and urban surveillance, problems which cannot endure poor prediction results and demand robust solutions.

\begin{figure}[t]
    \includegraphics[scale=0.300]{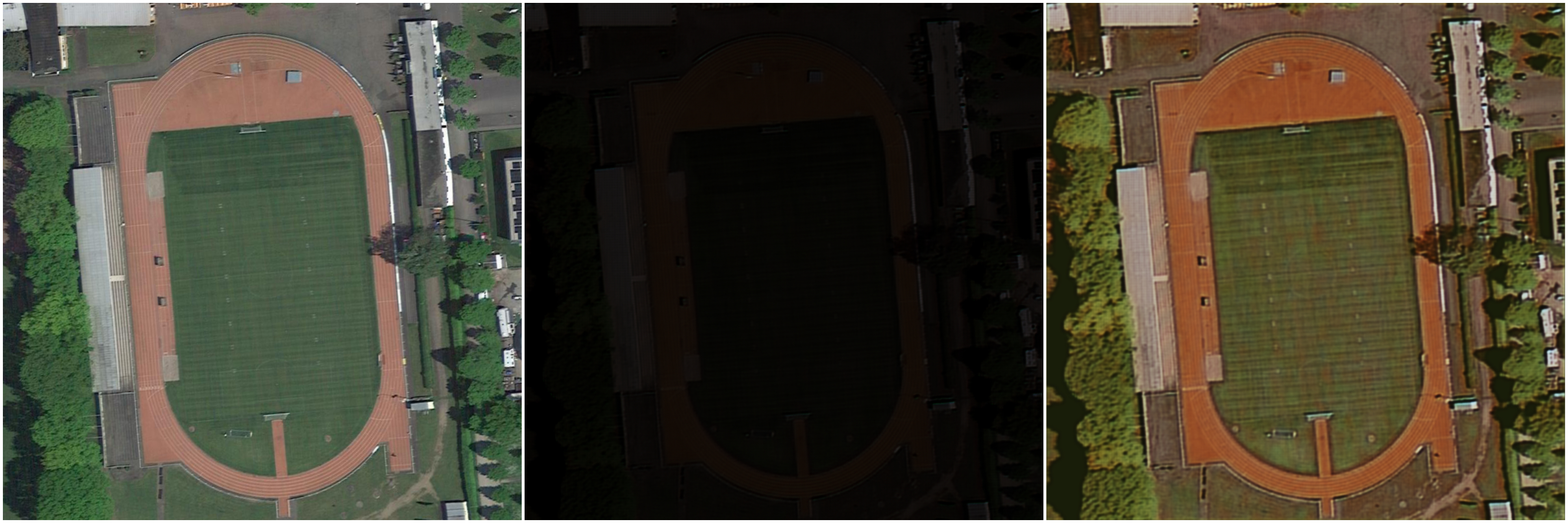}
    \caption{Low light aerial dataset generation. Left-original image; Middle-low light image generated using GANs; Right-enhanced low light image using light enhancement module. (Magnify for minor details in the images)} 
    \label{fig:ll-exm}
\end{figure}

In this study, we propose a novel network capable of performing simultaneous low light enhancement coupled with instance segmentation on the aerial images. The entire architecture is an end-to-end trainable CNN and delivers an increased quality of predictions. The overall contributions of this study are:
(1) We propose a self-supervised light enhancement network capable of performing instance segmentation on low light aerial images. 
(2) We generate a new low light aerial image dataset having annotations for instance segmentation.
(3) We demonstrate the effectiveness of our proposed network on the low light aerial dataset both qualitatively and quantitatively, thus laying the groundwork for future research.

\section{Dataset Generation}
To the best of our knowledge, no dataset provides low light aerial images along with annotations for instance segmentation. To validate our proposed method, we generate a low light aerial image dataset using iSAID \cite{waqas2019isaid}, LOw-Light dataset (LOL) \cite{Chen2018DeepRetinex} and GANs. The iSAID (outdoor, aerial) and LOL (indoor, terrestrial) datasets belong to different domains and lack one-to-one mapping between them. In order to efficiently translate the low light attribute from LOL to iSAID, we train them on the CycleGAN \cite{zhu2017unpaired} architecture, which is based on cycle consistency loss. We generate 18528 low light aerial images for the training set. All generated images are superior in quality and match with real-time low light conditions (Figure~\ref{fig:ll-exm}).

\begin{figure}[t]
    \includegraphics[width=\columnwidth]{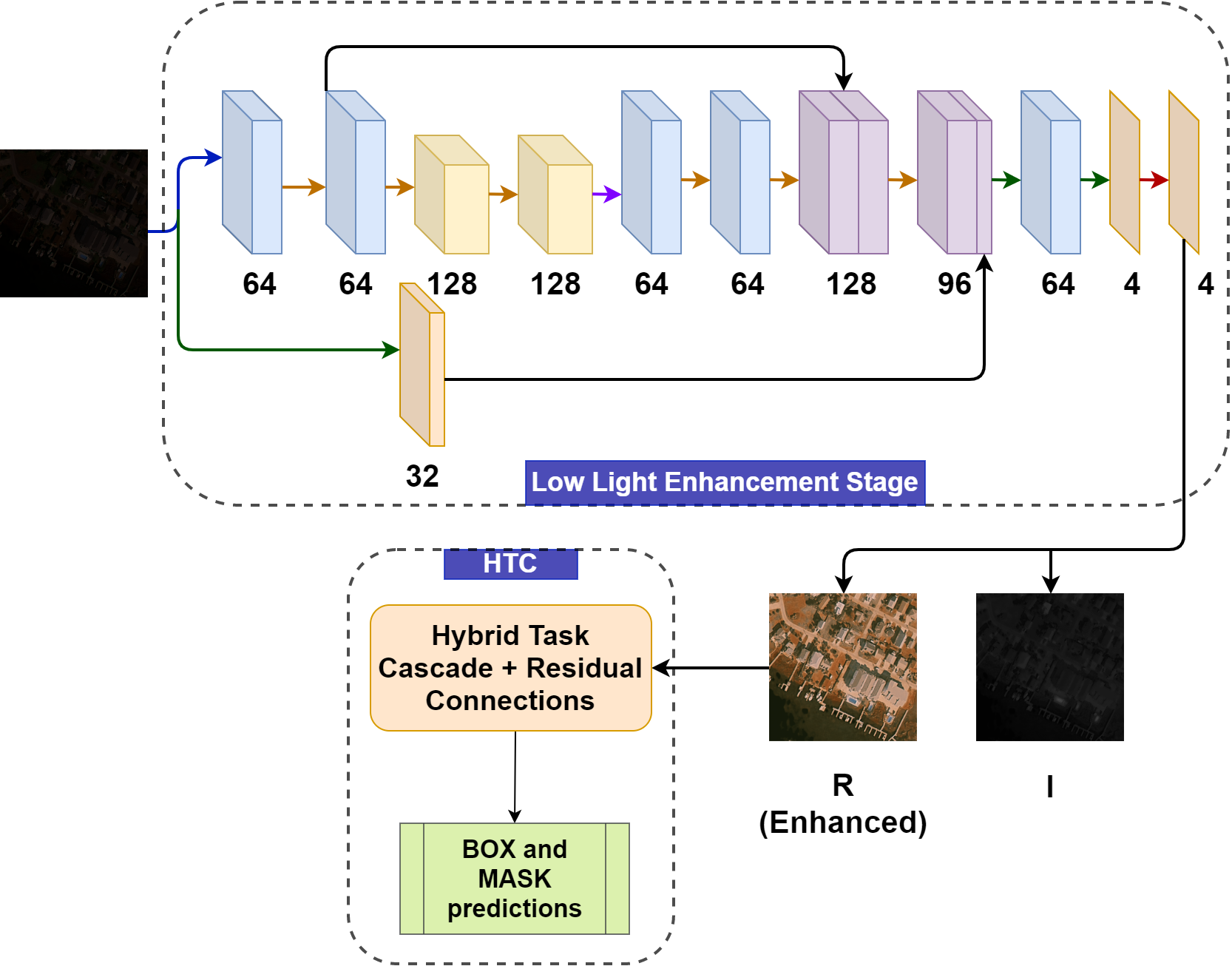}
    \caption{The architectural representation of the proposed
method. (Magnify for minor details in the images)} 
    \label{fig:ll-net}
\end{figure}

\section{Proposed Method}
The network is a two-phase architecture. The first phase deals with the brightness reconstruction of the image in a self-supervised manner, whereas the second phase is associated with the computer vision task of segmentation and detection.

\textbf{1. Self-Supervised Low Light Enhancement.} A self-supervised CNN module is used in this phase to eliminate the need of ground truth clear images or paired training data. This reduces the computational complexity of the network and makes the training process more generalisable. The method utilises the approach proposed by Zhang et al. \cite{zhang2020self} based on the Maximum Entropy-based Retinex Model. The light enhancement task is based on the assumption that the histogram distribution of the maximum channel of the enhanced image should validate and follow the histogram distribution of the maximum channel of the low light image after it has been histogram equalised. This attribute eliminates the requirement of paired examples, and thus, the loss is constructed using low light image tensors only. The following equation (Eq. \ref{eq1}) generalises the aforementioned concept.

\begin{dmath}
Z = \lVert  S - R \cdot I  \rVert + \lambda_1 \Big\lVert \smash{\displaystyle\max_{c \in R,G,B}} R^c - F( \smash{\displaystyle\max_{c \in R,G,B}} S^c ) \Big\rVert \\
+ \lambda_2 \lVert \Delta\ I \cdot \lambda exp(-\lambda_3 \Delta R)  \rVert + \lambda_4 \lVert \Delta R  \rVert
\label{eq1}
\end{dmath}

where S represents the low light image, R and I represent the reflectance and illuminance of the image, and $\lambda_i$ are the weight parameters. This composite equation can be solved by a deep learning network which can fragment the given image into illuminance and reflectance and employ this equation as its loss function.
The deep learning network utilised is shown in Figure~\ref{fig:ll-net}, which is a very elementary CNN structure apt for our usage, since it is fast, efficient and lightweight (2 MB). In the network, the upsampling and downsampling convolutions help in reducing noise but make the image tensors blurry. This flaw can be rectified by using extra convolutions post the sampling operations that can aid in the correct reconstruction of the target image.

\textbf{2. Instance Segmentation Network.} To perform instance segmentation on the enhanced image, we utilise Hybrid Task Cascade \cite{chen2019hybrid} along with some novel improvements that boost the prediction accuracy. We implement residual connection in the mask pipeline of HTC, which allows an efficient flow of the computed mask features in the pipeline. These connections help in retaining the mask features of tiny objects (which are abundant in aerial images) deduced in the initial stages throughout the final stages as well. 

The light enhancement module and HTC complement each other during the end-to-end training process, and losses from both phases contribute to the global loss of the joint network. We train the network for 11 epochs on the train set of the low light aerial dataset and obtain good results (Table~\ref{tab:arch-AP}) on the val set. Addition of a low light enhancing module considerably boosts the prediction scores of the instance segmentation network.

    \begin{table}[t]
    \def\arraystretch{1.4}
    \resizebox{\columnwidth}{!}{
        {\begin{tabular}{c|c|c c c|c c }
        \hline
       
        \textbf{Method} & \textbf{Type} & \textbf{AP} & \textbf{AP\textsubscript{50}} & \textbf{AP\textsubscript{75}} & \textbf{AP\textsubscript{S}} & \textbf{AP\textsubscript{M}} \\
        \hline
        \hline
            \multirow{2}{*}{Original HTC} & Box & 38 & 55.9 & 42.1 & 21.1 & 45.9 \\
            \cline{2-7}
            & Segm & 31.3 & 51.9 & 32.7 & 16.1 & 37.8 \\
            \hline
            \multirow{2}{*}{\textbf{Proposed Method}} & Box & 38.9 & 57.5 & 43.1 & 21.7 & 47.2 \\
            \cline{2-7}
            & Segm & 32.2 & 53.8 & 33.9 & 17 & 39.2 \\
        \hline
        
        \end{tabular}}\quad
        }
        \caption{Performance on the val set of the low light aerial dataset.}
        \label{tab:arch-AP}
    \end{table}
    
\section{Conclusion}
In this abstract, we propose a new method for robust feature extraction from low light aerial images. Our method efficiently restores the brightness in the image and then performs instance segmentation on the enhanced image. The light enhancement module, adept in brightness reconstruction, is self-supervised and incurs a very minor computational cost. The instance segmentation algorithm is further improved by using residual connections, which aid in dense information flow in the network. In addition, we also generate a synthetic dataset consisting of low light aerial images and annotations for performing instance segmentation in adverse conditions. 
\section{Supplementary Material}
\subsection{Training Configuration}
\begin{figure}[t]
    \centering
    \includegraphics[width=\columnwidth]{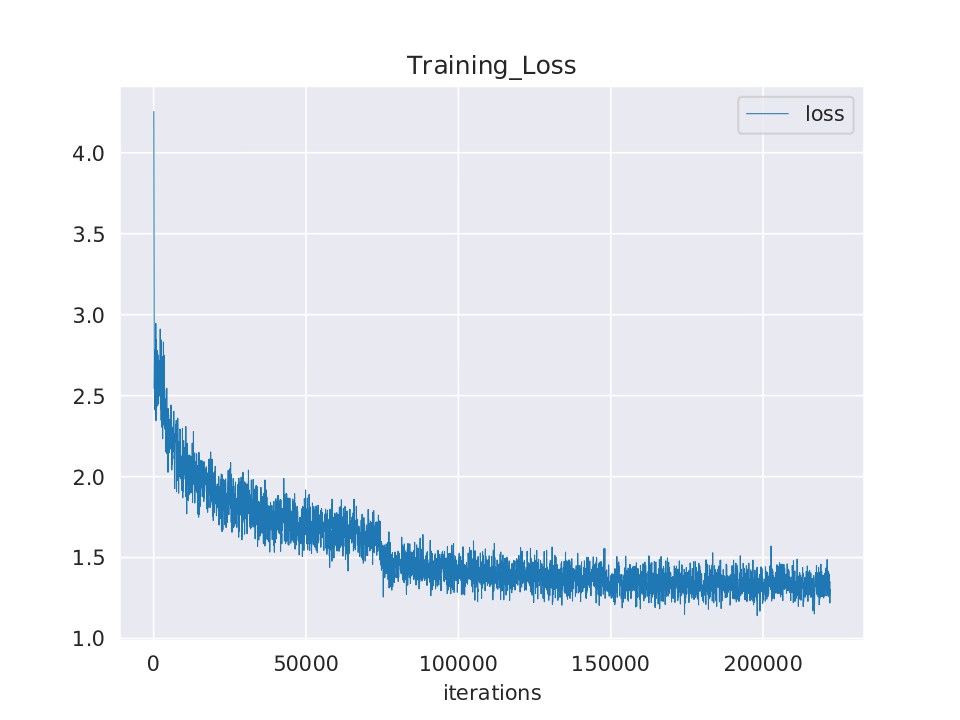}
    \caption{Training Loss for the complete network} 
    \label{fig:loss}
\end{figure}

\begin{figure}[t]
    \centering
    \includegraphics[width=\columnwidth]{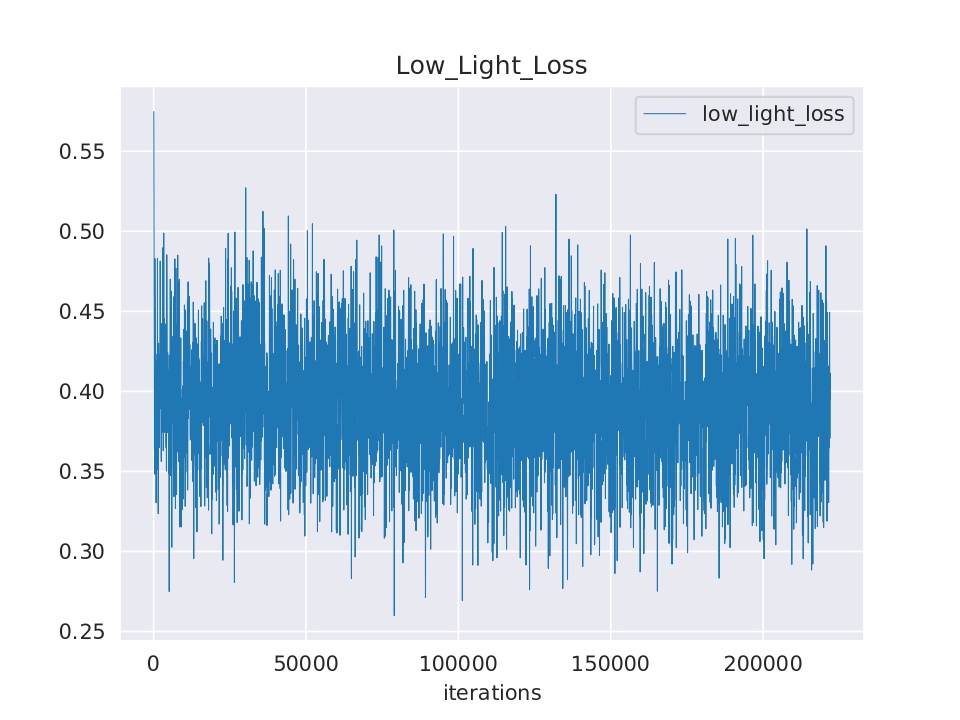}
    \caption{Low light loss}
    \label{fig:loss-ill}
\end{figure}

We train the method for 11 epochs at a batch size of 1 and a learning rate of 0.0031. The momentum and weight decay values are set to 0.9 and 0.0001 respectively. The learning rate is decayed by 10 times at epochs 4, 8 and 10. We adopt multi-scaled training approach where the shorter edge is randomly sampled from six different scales of 1200,1000,800,600 and 400. The network is trained on an NVIDIA Tesla V100 GPU, 32 GB memory. Following are the training trajectories for the (1) complete network (Fig~\ref{fig:loss}), and (2) the light enhancement module (Fig~\ref{fig:loss-ill}).

\subsection{Additional Quantitative Results}
Apart from the main results mentioned in the paper, we made some more fine observations that uphold the fact that low light conditions can hinder the performance of computer vision techniques substantially. We use HTC as our principal instance segmentation network and compare its performance on the val set of low light aerial dataset with two variants: (1) when it is trained with only normal images and (2) when it is trained with low light aerial images. It is evident from Table~\ref{tab:dataper} that the performance on low light images by the first approach is very poor and goes to as low as 17.5 and 15.2 APs in box and mask criteria. However, the performance of the second approach is quite exceptional, and the results obtained are almost double of those obtained in the first approach. This clearly shows that networks which are trained on clean and sanitised data do not perform well in adverse low light conditions.

\begin{table}[t]
    \renewcommand{\arraystretch}{1.4}
    \resizebox{\columnwidth}{!}{
        {\begin{tabular}{c|c|c c c|c c }
        \hline
       
        \textbf{Method} & \textbf{Type} & \textbf{AP} & \textbf{AP\textsubscript{50}} & \textbf{AP\textsubscript{75}} & \textbf{AP\textsubscript{S}} & \textbf{AP\textsubscript{M}} \\
        \hline
        \hline
            
            \multirow{2}{*}{HTC trained normally} & Box & 17.5 & 25.5 & 19.8 & 8.7 & 22.8 \\
            \cline{2-7}
            & Segm & 15.2 & 24.7 & 15.9 & 6.9 & 19.9 \\
        \hline
        \multirow{2}{*}{\textbf{HTC trained in low light}} & Box & 38 & 55.9 & 42.1 & 21.1 & 45.9 \\
            \cline{2-7}
            & Segm & 31.3 & 51.9 & 32.7 & 16.1 & 37.8 \\
            \hline
        
        \end{tabular}}\quad
        }
        \caption{Performance of HTC on the val set of the low light aerial dataset. In the first case, HTC is trained with normal images whereas it is trained with low light images in the second case.}
        \label{tab:dataper}
\end{table}

\begin{table}[t]
    \renewcommand{\arraystretch}{1.4}
    \resizebox{\columnwidth}{!}{
        {\begin{tabular}{c|c|c c c|c c }
        \hline
       
        \textbf{Method} & \textbf{Type} & \textbf{AP} & \textbf{AP\textsubscript{50}} & \textbf{AP\textsubscript{75}} & \textbf{AP\textsubscript{S}} & \textbf{AP\textsubscript{M}} \\
        \hline
        \hline
            \multirow{2}{*}{HTC trained in low light} & Box & 41.7 & 61.8 & 46 & 25.2 & 49.3 \\
            \cline{2-7}
            & Segm & 34.3 & 57.6 & 35.8 & 19.4 & 41.1 \\
            \hline
        
        \end{tabular}}\quad
        }
        \caption{Performance of HTC on the val set of iSAID dataset, when it is trained with low light images}
        \label{tab:arch-AP}
    \end{table}

We also evaluate the performance of the second variant of HTC on clear aerial images. Table~\ref{tab:arch-AP} clearly demonstrates that even though HTC was trained on low light aerial images, it maintains good generalisation and performs significantly well on clear normal images too. This observation indicates that training networks with dataset having adverse conditions provides them a good generalising power, whereas not doing so may lead to poor performance in disadvantageous scenarios.

\subsection{Dataset Generation: Additional Details}
\begin{figure*}[t]
    \centering
    \includegraphics[scale=0.7]{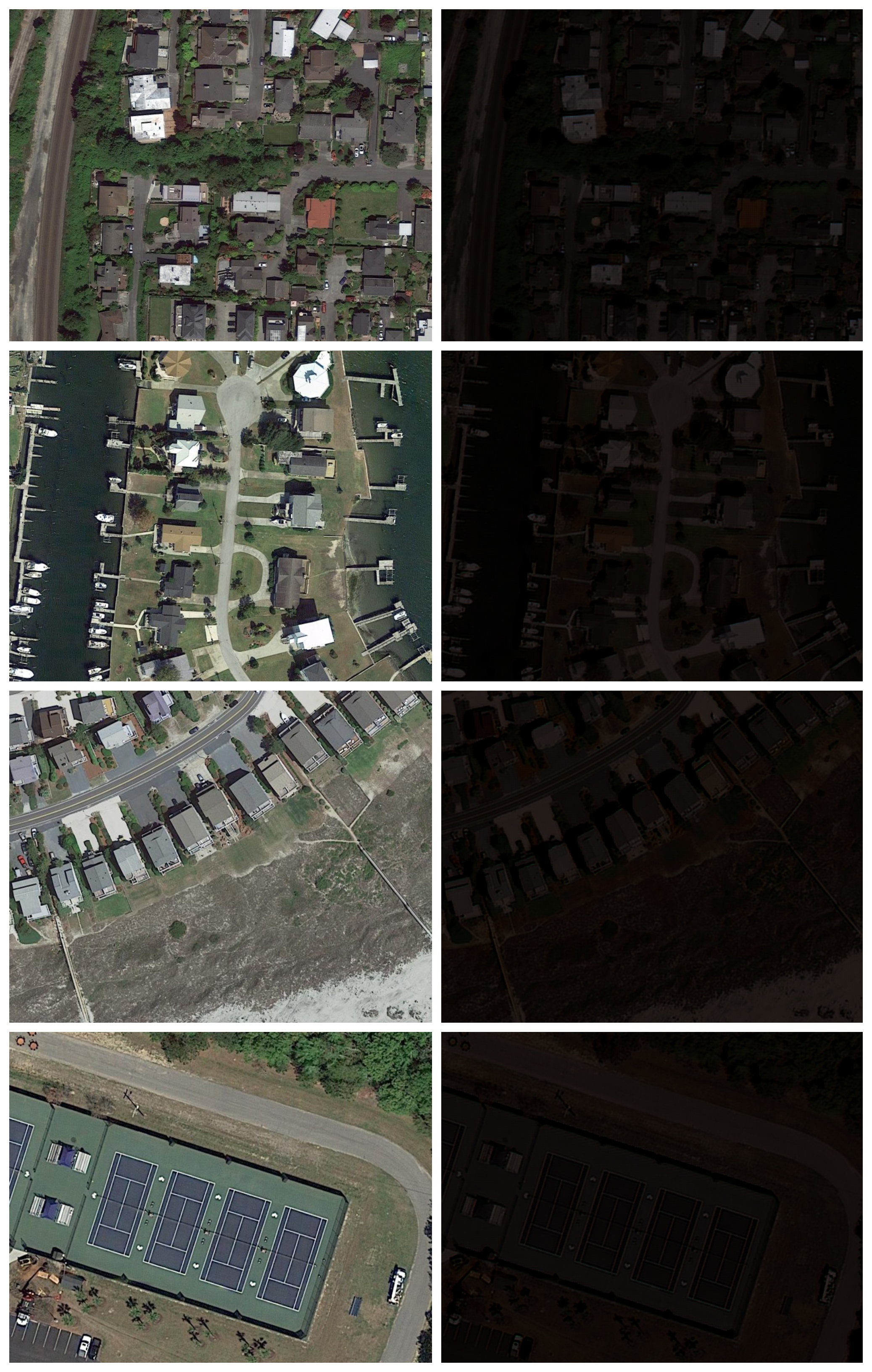}
    \caption{Low light dataset generated from iSAID and LOL using CycleGAN. Images on the left correspond to the iSAID dataset, images on the right are synthetically generated.}
    \label{fig:ll-data}
\end{figure*}

We generate a new low light aerial image dataset, consisting of mask and box annotations for instance segmentation using a GAN based approach. We generate this by transferring the low light attribute from the LOL dataset to iSAID, which is an aerial dataset consisting of annotations in COCO format. We use a CycleGAN for this purpose and train it for 100 epochs at a learning rate of 0.0001. The GAN mode is set to 'lsgan' configuration whereas the discriminator mode is configured with a 'basic' setting. We set the crop size to 600 during the training process in order to maximize the field view of the GAN model. Fig~\ref{fig:ll-data} represents some examples of the low light images generated. It is clear that the generated images successfully capture the low light property and retain the core feature representation of the original image.

\subsection{Testing the Low Light Enhancement Module}
\begin{figure*}[t]
    \centering
    \includegraphics[scale=0.8]{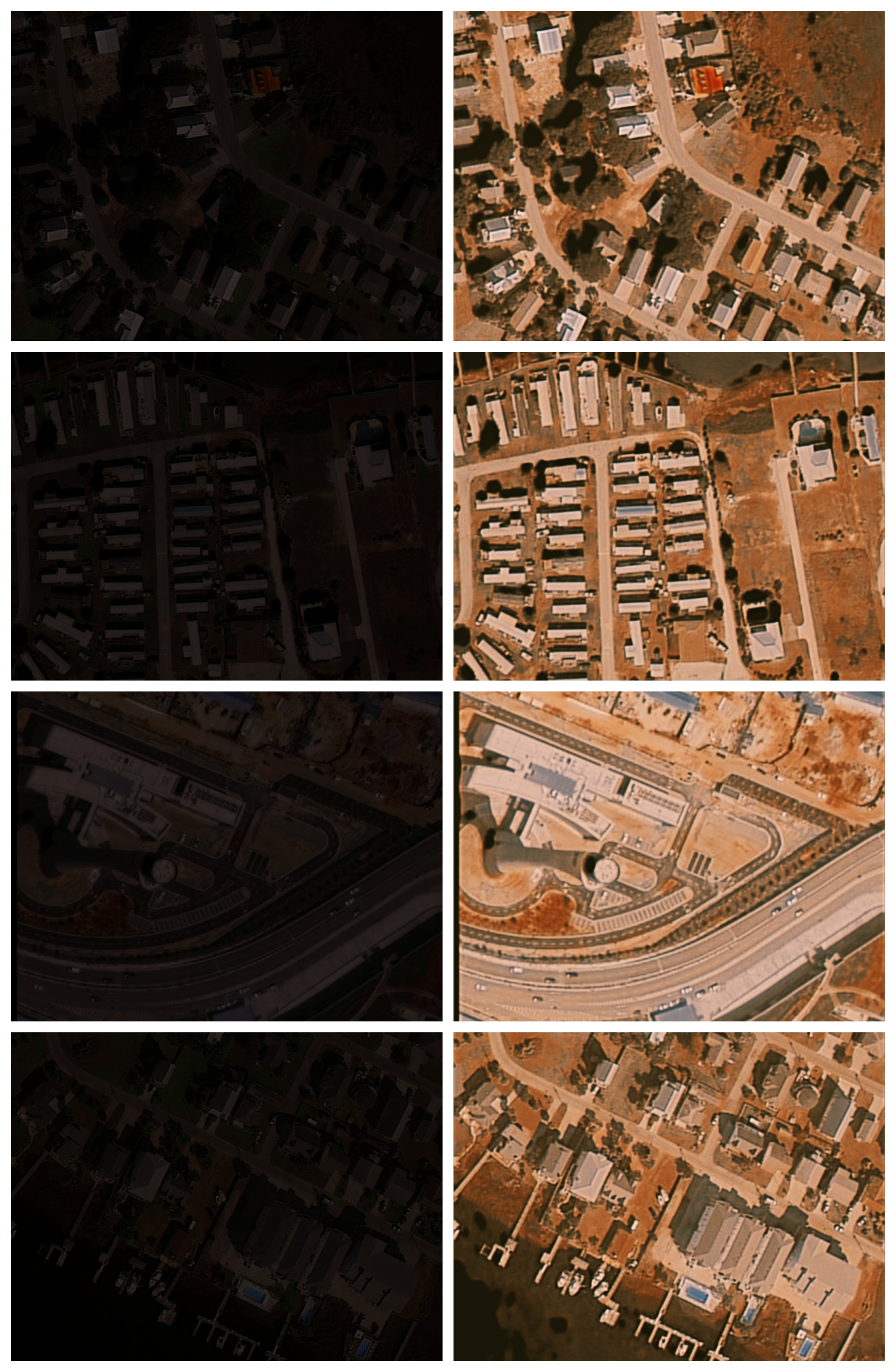}
    \caption{Enhancement of low light images using the self-supervised light enhancement module.}
    \label{fig:llrecov}
\end{figure*}

We train and validate the performance of the self-supervised low light enhancement module separately from the main method proposed in our work. We train the CNN module for 85 epochs on the low light aerial dataset at a learning rate of 0.0001. At the end of the training, we test the performance of the trained module on some dark test images.
Fig~\ref{fig:llrecov} illustrates that the light enhancement module is proficient in restoring the brightness of the image. All the core features in the image have been retained and the image has been enhanced which facilitates good performance of the computer vision techniques.

\section{Acknowledgements}
This work is supported by BITS Additional Competitive
Research Grant (PLN/AD/2018-19/5).

\small{
\bibliography{references.bib}
}

\end{document}